\documentclass[sigconf]{aamas} 

\usepackage{balance}
\usepackage{subcaption}
\usepackage{balance} 
\usepackage{amsmath}
\usepackage{algorithm}
\usepackage[noend]{algpseudocode}
\usepackage{tikz}
\usetikzlibrary{calc}

\DeclareMathOperator*{\E}{\mathbb{E}}
\DeclareMathOperator*{\argmax}{arg\,max}

\makeatletter
\def\BState{\State\hskip-\ALG@thistlm}
\makeatother

\setcopyright{ifaamas}
\acmConference[AAMAS '23]{Proc.\@ of the 22nd International Conference
on Autonomous Agents and Multiagent Systems (AAMAS 2023)}{May 29 -- June 2, 2023}
{London, United Kingdom}{A.~Ricci, W.~Yeoh, N.~Agmon, B.~An (eds.)}
\copyrightyear{2023}
\acmYear{2023}
\acmDOI{}
\acmPrice{}
\acmISBN{}

\acmSubmissionID{1032}

\title[AAMAS-2023 Paper]{The Benefits of Power Regularization in Cooperative Reinforcement Learning}

\author{Michelle Li}
\affiliation{
  \institution{Massachusetts Institute of Technology}
  \city{Cambridge, MA}
  \country{}}
\email{michelleli@alum.mit.edu}

\author{Michael Dennis}
\affiliation{
  \institution{University of California, Berkeley}
  \city{Berkeley, CA}
  \country{}}
\email{michael_dennis@cs.berkeley.edu}

\begin{abstract}
Cooperative Multi-Agent Reinforcement Learning (MARL) algorithms, trained only to optimize task reward, can lead to a concentration of power where the failure or adversarial intent of a single agent could decimate the reward of every agent in the system.  In the context of teams of people, it is often useful to explicitly consider how power is distributed to ensure no person becomes a single point of failure. Here, we argue that explicitly regularizing the concentration of power in cooperative RL systems can result in systems which are more robust to single agent failure, adversarial attacks, and incentive changes of co-players.  To this end, we define a practical pairwise measure of power that captures the ability of any co-player to influence the ego agent's reward, and then propose a power-regularized objective which balances task reward and power concentration.  Given this new objective, we show that there always exists an equilibrium where every agent is playing a power-regularized best-response balancing power and task reward. Moreover, we present two algorithms for training agents towards this power-regularized objective: Sample Based Power Regularization (SBPR), which injects adversarial data during training; and Power Regularization via Intrinsic Motivation (PRIM), which adds an intrinsic motivation to regulate power to the training objective. Our experiments demonstrate that both algorithms successfully balance task reward and power, leading to lower power behavior than the baseline of task-only reward and avoid catastrophic events in case an agent in the system goes off-policy.

\end{abstract}


\newcommand{\BibTeX}{\rm B\kern-.05em{\sc i\kern-.025em b}\kern-.08em\TeX}

\begin{document}


\pagestyle{fancy}
\fancyhead{}


\maketitle

\section{Introduction}
\newcommand{\payoff}[4][below]{\node[#1]at(#2){$(#3,#4)$};}
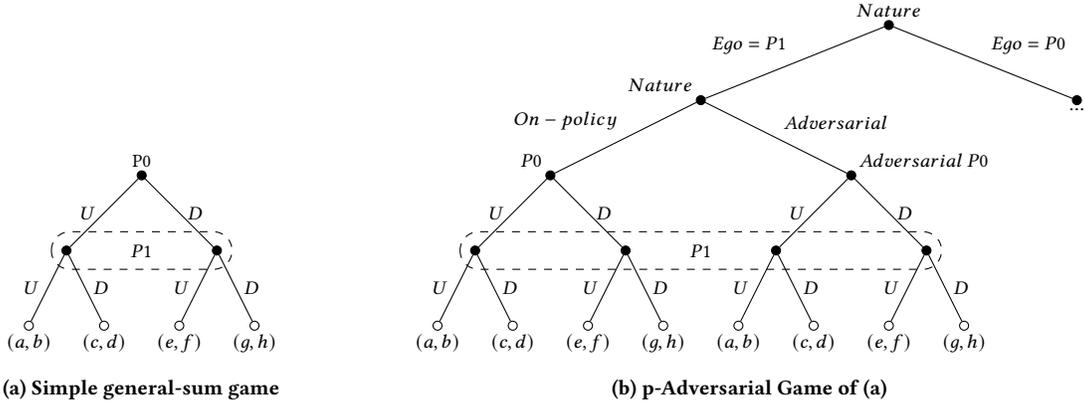
\begin{figure*}
\begin{subfigure}{.3\textwidth}
\centering
\begin{tikzpicture}[scale=1,font=\footnotesize]
\centering
\tikzstyle{solid node}=[circle,draw,inner sep=1.2,fill=black];
\tikzstyle{hollow node}=[circle,draw,inner sep=1.2];
\tikzstyle{level 1}=[level distance=10mm,sibling distance=20mm]
\tikzstyle{level 2}=[level distance=10mm,sibling distance=10mm]

\node(1)[solid node]{}
child{node[solid node]{}
child{node[hollow node]{}edge from parent node[left]{$U$}}
child{node[hollow node]{}edge from parent node[right]{$D$}}
edge from parent node[left]{$U$}
}
child{node[solid node]{}
child{node[hollow node]{}edge from parent node[left]{$U$}}
child{node[hollow node]{}edge from parent node[right]{$D$}}
edge from parent node[right]{$D$}
};
\draw[dashed,rounded corners=7]($(1-1)+(-.2,.25)$)rectangle($(1-2)+(.2,-.25)$);

\node[above]{P0};
\node at ($.5*(1-1)+.5*(1-2)$) {$P1$};
\payoff{1-1-1}ab
\payoff{1-1-2}cd
\payoff{1-2-1}ef
\payoff{1-2-2}gh
\end{tikzpicture}
\caption{Simple general-sum game}
\label{fig:efg_normal_game}
\end{subfigure}
\begin{subfigure}{.6\textwidth}
\centering
\begin{tikzpicture}[scale=1,font=\footnotesize]
\centering
\tikzstyle{solid node}=[circle,draw,inner sep=1.2,fill=black];
\tikzstyle{hollow node}=[circle,draw,inner sep=1.2];
\tikzstyle{level 1}=[level distance=10mm,sibling distance=50mm]
\tikzstyle{level 2}=[level distance=10mm,sibling distance=40mm]
\tikzstyle{level 3}=[level distance=10mm,sibling distance=20mm]
\tikzstyle{level 4}=[level distance=10mm,sibling distance=10mm]

\node(1)[solid node]{}
child{node(2)[solid node]{}
child{node(3)[solid node]{}
child{node[solid node]{}
child{node[hollow node]{}edge from parent node[left]{$U$}}
child{node[hollow node]{}edge from parent node[right]{$D$}}
edge from parent node[left]{$U$}
}
child{node[solid node]{}
child{node[hollow node]{}edge from parent node[left]{$U$}}
child{node[hollow node]{}edge from parent node[right]{$D$}}
edge from parent node[right]{$D$}
}
edge from parent node[above left]{$On-policy$}
}
child{node(4)[solid node]{}
child{node[solid node]{}
child{node[hollow node]{}edge from parent node[left]{$U$}}
child{node[hollow node]{}edge from parent node[right]{$D$}}
edge from parent node[left]{$U$}
}
child{node[solid node]{}
child{node[hollow node]{}edge from parent node[left]{$U$}}
child{node[hollow node]{}edge from parent node[right]{$D$}}
edge from parent node[right]{$D$}
}
edge from parent node[above right]{$Adversarial$}
}
edge from parent node[above left]{$Ego=P1$}
}
child{node[solid node]{}
edge from parent node[above right]{$Ego=P0$}
};
\draw[dashed,rounded corners=7]($(3-1)+(-.2,.25)$)rectangle($(4-2)+(.2,-.25)$);

\node[above]at(1){$Nature$};
\node[above left]at(2){$Nature$};
\node[below]at(1-2){$...$};
\node at ($.5*(3-1)+.5*(4-2)$) {$P1$};
\node[above left]at(3){$P0$};
\node[above right]at(4){$Adversarial$ $P0$};
\payoff{3-1-1}ab
\payoff{3-1-2}cd
\payoff{3-2-1}ef
\payoff{3-2-2}gh
\payoff{4-1-1}ab
\payoff{4-1-2}cd
\payoff{4-2-1}ef
\payoff{4-2-2}gh
\end{tikzpicture}
\caption{p-Adversarial Game of (a)}
\label{fig:efg_for_existence_proof}
\end{subfigure}
\caption{Extensive Form diagrams of a simple 1 timestep game and its corresponding p-Adversarial game at s where Nature chooses an ego agent and whether to use the on-policy or adversarial co-player toward the ego agent. The dashed lines encircle nodes in the same information set for Player 1 because agents act simultaneously. We omit a final layer of Nature nodes that model the probabilistic transition function.}
\end{figure*}

When considering how to optimally structure a team, institution, or society, a key question is how responsibility, power, and blame ought to be distributed, and as such it is a broadly studied concept in the social sciences~\cite{lukes2021power,mann2012sources}. We often want to avoid too much power lying in the hands of a few actors, preferring power to be distributed rather than concentrated. One example of a distributed system of power is the US government: in principle, having three branches with checks and balances helps prevent any single branch from having too much power.

The same basic idea applies in Multi-Agent Reinforcement Learning (MARL) systems: regardless of whether the setting is fully cooperative, fully competitive, or general sum, it is often advantageous for agents to limit the amount of power other agents have. 

Power resists formalization despite being a prevalent and intuitive concept. In this paper, we make no normative claims about how power ought to be defined or regulated, just that power is a conceptually useful tool for cooperative multi agent systems. To focus on empirical progress, we limit our attention to power as \textit{influence on reward}.  But regardless of how it is formalized, we argue that avoiding concentration of power can simultaneously mitigate three problems that are generally thought of separately: system failure, adversarial attacks, and incentive changes.  It does so by mitigating the effects of off policy behavior, regardless of the cause.

Consider the following toy example: a group of agents need to work cooperatively to maximize production. Agents produce output by starting from raw material and applying a series of $m$ actions. They can either work individually or form an assembly line, which is more efficient because of specialization and batch productivity but requires that every agent is a single point of failure.  Changes in any agent's behavior would bring the whole system to a halt. Depending on how much we care about task reward versus robustness, we might prefer one behavior or the other.


Our contributions in this work are as follows: 1) We propose a practical measure of power amenable to optimization -- how much another agent can decrease our return by changing their action for one timestep. 2) We propose a framework for balancing maximizing task reward and minimizing power by regularizing the task objective for power, and then show an equilibrium always exists with this modified objective.  3) We present two algorithms for achieving power regularization: one, Sample Based Power Regularization (SBPR), which injects adversarial data during training by adversarially perturbing one agent's actions with some probability at any timestep; and two, Power Regularization via Intrinsic Motivation (PRIM), which adds an intrinsic reward to regularize power at each timestep. SBPR is simpler but PRIM is better able to achieve the right reward-power tradeoffs for very small values of $\lambda$. Our experiments in an Overcooked-inspired environment~\cite{carroll2019utility} demonstrate that both algorithms can achieve various power-reward tradeoffs and can reduce power compared to the task reward-only baseline. 

\section{Related Work}
Power has been broadly studied in the social sciences~\cite{lukes2021power,mann2012sources} and for multi-agent systems, but has not been studied in the context of deep multi-agent RL to the best of our knowledge.  Instead, prior work has focused on graph-theoretic analysis~\cite{grossi2006structural} or symbolic formulations~\cite{dignum2006coordinating}.  There have also been productive formulations of the related concepts of responsibility and blame~\cite{chockler2004responsibility,gerstenberg2015responsibility,alechina2020causality,halpern2018towards,friedenberg2019blameworthiness}, which have strong connections to power.

In AI, power has been formalized in a single agent context, with recent work towards defining power and regularizing an agent's own behavior with respect to power~\cite{turner2019optimal,turner2020avoiding,turner2020conservative}.  While it is a promising direction to extend these formal measures to MARL, we focus on making empirical progress on regularizing power in this work.

Though the literature on power in deep MARL is sparse, the literature on the problems that power regulation can help mitigate is more robust.  For instance, there is a large body of work on designing MARL systems that cooperate robustly in sequential social dilemmas~\cite{leibo2017multi,foerster2017learning,hughes2018inequity,jaques2019social}, in which balancing power amongst the agents is critical.  There is also a significant body of work showing the existence of adversarial attacks for neural networks~\cite{szegedy2013intriguing} and single- and multi-agent RL~\cite{huang2017adversarial,kos2017delving,lin2017tactics,gleave2019adversarial}.  In such cases, power regularization can help build fault tolerance.

Finally, the algorithms we present for power regularization may be well suited for situations where one desires agents that work well with unknown teammates, if combined with approaches from ad-hoc teamwork~\cite{AAAI10-adhoc,barrett2011empirical} and zero-shot coordination~\cite{hu2020other,treutlein2021new}.  This is especially true in situations where agents must infer who to trust (i.e. who to entrust power to)~\cite{littman2001friend,serrino2019finding}.



\section{Background}
We model our setting as a Markov game~\cite{littman1994markov} defined by the tuple $(N,S,A,T,R, \gamma)$ where $N$ is the number of players, $S$ is the set of environment states, $A = \times_{i \in \{0,\dotsc, N\}} A_i$ is the joint action space, $T: S \times A \rightarrow S$ is the transition function, $R = \times_{i \in \{0,\dotsc, N\}} R_i$ is the reward function for each player, and $\gamma \in (0,1]$ is the discount factor.  At every timestep, each agent $i$ chooses an action $a_i \in A_i$, and the joint action $a=(a_1,a_2,...,a_n) \in A$ is used to transition the environment according to $T(s'|a,s)$. Each agent $i$ receives their own reward $r_i=R_i(s,a)$. While our theoretical results hold for general-sum Markov games, we only empirically evaluate on fully cooperative environments, thus assuming that all agents have the same reward: $R_i=R_j\,\forall i,j$. We operate in the fully-observed setting: each agent can directly observe the true state $s$. Each agent $i$ aims to independently maximize their time-discounted, expected reward $U_i = \E[\sum_{t=0}^T \gamma^t r_t]$ where $r_t$ is the reward at time $t$. Throughout, we will assume finite, discrete actions and finite time. 

In any game, it is useful for a player to consider their \emph{best responses}: optimal policies given fixed policies for the co-players:
\[
\pi_i \in \argmax_{\pi_i' \in \Pi_i}\{U_i(\pi_i'; \pi_{-i})\} = BR(\pi_{-i}).
\]
where $\pi_{-i}$ refers to the policies for all players other than $i$.  Furthermore, we say that a policy $\pi_i$ is a \textit{local best response} at state $s$ if it chooses an optimal distribution of actions at $s$ given the rest of its policy $\pi_i$ and co-player policies $\pi_{-i}$.  That is, 
\[
\pi_i \in \argmax_{\pi_i' \in \overline{\Pi}_i(\pi_i;s)}\{U_i(\pi_i'; \pi_{-i})\} = BR^{local}(\pi_{-i}).
\]
where $\overline{\Pi}_i(\pi_i;s) = \{\pi'_i \in \Pi_i \vert \forall s' \neq s, \pi'_i(s') = \pi(s')\}$

A set of policies $\pi$ for each player where each policy $\pi_i$ is a best response to the other strategies is called a Nash equilibrium that is, $\forall i, \pi_i \in BR(\pi_{-i}))$. This is a "stable point" where no agent has an incentive to unilaterally change their policy. Furthermore, we say that a set of policies $\pi$ form a \textit{local Nash equilibrium} at state $s$ if all policies are a local best-response to the other policies at $s$.


\section{Formalism} \label{sec:formalism}
To design systems that regularize for power, it is important to be clear about our objective and how we define power.

\subsection{Measuring Power}

Our main goal is to make empirical progress on building power-regularizing MARL systems, so we will not aim to find the most proper or most general definition of power. We define a measure for power of co-player $j$ over ego agent $i$, which we call 1-step adversarial power, as the difference $j$ could make on $i$'s reward if $j$ had instead acted adversarially toward $i$ for one timestep.

\begin{definition}[1-step adversarial power] \label{def:power}
Let $U_i^{\text{task}}$ denote player $i$'s task utility. Given policies $\pi$, the 1-step adversarial power agent $j$ has on agent $i$ when starting from state $s$ is:
\[\texttt{power}(i, j \vert s, \pi) = r + \E[U_i^{\text{task}}(s', \pi)] - \min_{a_j \in A_j }(r_{a_j}+\E[U_i^{\text{task}}(s_{a_j}', \pi)])\]
where $s' = T(s,\pi)$, $s_{a_j}' = T(s,a_j;\pi_{-j})$, and $r$ and $r_{a_j}$ are $i$'s rewards obtained on-policy and with $j$'s deviation to $a_j$, respectively.
\end{definition}


The $\min_{a_j \in A_j}$ is taken over the set of deterministic actions. Note that it is not necessary to consider stochastic policies as the most powerful stochastic $\pi_j$ could simply place probability $1$ on any of the deterministic actions that achieve the lowest utility for $i$.



Counterintuitively, it is possible that all of $j$'s immediate actions exert some causal effect on $i$'s utility without $j$ having any power over $i$. This can happen if all of $j$'s actions reduce $i$'s reward by $10$, for example. Since we are defining power in relative terms, if all actions have the same effect on $i$, we say $j$ has no power over $i$ as any causal effects of $j$'s actions on $i$ are inevitable. See Table~\ref{Tab:AllEquallyBadForB} for an example worked out explicitly.  Such nuance is reminiscent of the difficulties in defining blame~\cite{chockler2004responsibility}.  Exploring such connections in-depth could be a path towards better metrics for measuring power. 

\begin{table}
\centering
    \caption{An example game where player 1 has no power over player 2 because all of 1's actions are equally bad for 2.}
    \begin{tabular}{cc|c|c|c|}
      & \multicolumn{1}{c}{} & \multicolumn{1}{c}{$X$}  & \multicolumn{1}{c}{$Y$}  & \multicolumn{1}{c}{$Z$} \\\cline{3-5}
                & $X$ & $(3,-10)$ & $(3,-9)$ & $(3,-8)$ \\\cline{3-5}
                & $Y$ & $(2,-10)$ & $(2,-9)$ & $(2,-8)$ \\\cline{3-5}
                & $Z$ & $(1,-10)$ & $(1,-9)$ & $(1,-8)$ \\\cline{3-5}
    \end{tabular}
    \label{Tab:AllEquallyBadForB}
\end{table}

\subsection{Regularizing for Power}
\label{section:regularizing-for-power}
Traditionally, cooperative MARL algorithms aim to optimize the discounted sum of task rewards, which we call \emph{task utility}, without explicit consideration for the amount of power held by other agents. We argue that we can make systems more robust by optimizing an explicit trade-off between maximizing task reward and minimizing power. This framework has the advantage of addressing system failure, adversarial attacks, and incentive changes all at once by mitigating the negative effects of off-policy behavior.

We focus on linear trade-offs, that is objectives of the form
\begin{equation} \label{eq:regularized_objective}
    U_i(\pi \vert s)=U_i^{\text{task}}(\pi \vert s) +\lambda U_i^{\text{power}}(\pi \vert s)›
\end{equation}

where $U_i^{\text{task}}(\pi \vert s)$ is the task utility for player $i$ starting in state $s$ and $U_i^{\text{power}}(\pi \vert s) = \sum_{t=0}^T R_i^{power}(s_t, \pi)$ is the sum of power rewards $R_i^{power}$ at states starting from $s$ reached by unrolling $\pi$. In the $2$-agent setting, $R_i^{power}(s, \pi)=-\texttt{power}(i,j \vert s, \pi)$, but with more agents, $R_{power}$ must aggregate information about the powers of all $j$ on $i$. In our experiments with more than $2$ agents we let $R_i^{power}(s, \pi)=-\frac{1}{N-1}\sum_j \texttt{power}(i,j \vert s, \pi)$ (the mean function). We leave the problem of determining the most appropriate choice of aggregation function to future work.

Consider the $2$-player general-sum matrix game defined in Table ~\ref{Tab:AttackDefenseGame} which we call the Attack-Defense game. This is a single timestep game so $U_i^{\text{power}}(\pi \vert s)=-\texttt{power}(i,j \vert s, \pi)$. If either player plays $X$, the other player could play $Z$ reducing the utility. playing $Y$ to guarantees $2$ utility, paying a small price to reduce power. 

\begin{table*}
\begin{minipage}{.46\textwidth}
\centering
\caption{The Attack-Defense Game: an opponent can take away your utility if you play $X$, but you can pay a small cost to defend against that by playing $Y$.}
\label{Tab:AttackDefenseGame}
\begin{tabular}{cc|c|c|c|}
  & \multicolumn{1}{c}{} & \multicolumn{1}{c}{$X$}  & \multicolumn{1}{c}{$Y$}  & \multicolumn{1}{c}{$Z$} \\\cline{3-5}
      & $X$ & $(3,3)$ & $(3,2)$ & $(0,0)$ \\\cline{3-5}
      & $Y$ & $(2,3)$ & $(2,2)$ & $(2,0)$ \\\cline{3-5}
      & $Z$ & $(0,0)$ & $(0,2)$ & $(0,0)$ \\\cline{3-5}
\end{tabular}
\end{minipage}
\begin{minipage}{.5\textwidth}
\centering
\caption{The Larger Attack-Defense Game.}
\label{Tab:LargerAttackDefenseGame}
\begin{tabular}{cc|c|c|c|c|c|c|}
  & \multicolumn{1}{c}{} & \multicolumn{1}{c}{$A$}  & \multicolumn{1}{c}{$B$}  & \multicolumn{1}{c}{$C$}  & \multicolumn{1}{c}{$D$}  & \multicolumn{1}{c}{$E$}  & \multicolumn{1}{c}{$F$} \\\cline{3-8}
      & $A$ & $(3,3)$ & $(3,2.5)$ & $(3,2)$ & $(3,1.5)$ & $(3,1)$ & $(-2,0)$ \\\cline{3-8}
      & $B$ & $(2.5,3)$ & $(2.5,2.5)$ & $(2.5,2)$ & $(2.5,1.5)$ & $(2.5,1)$ & $(0,0)$ \\\cline{3-8}
      & $C$ & $(2,3)$ & $(2,2.5)$ & $(2,2)$ & $(2,1.5)$ & $(2,1)$ & $(0.75,0)$ \\\cline{3-8}
      & $D$ & $(1.5,3)$ & $(1.5,2.5)$ & $(1.5,2)$ & $(1.5,1.5)$ & $(1.5,1)$ & $(1,0)$ \\\cline{3-8}
      & $E$ & $(1,3)$ & $(1,2.5)$ & $(1,2)$ & $(1,1.5)$ & $(1,1)$ & $(1,0)$ \\\cline{3-8}
      & $F$ & $(0,-2)$ & $(0,0)$ & $(0,0.75)$ & $(0,1)$ & $(0,1)$ & $(0,0)$ \\\cline{3-8}
\end{tabular}
\end{minipage}
\end{table*}

If optimizing purely for task reward, both agents play $X$ to achieve the utility-maximizing Nash equilibrium. However, playing $X$ incurs $3$ power while playing $Y$ incurs $0$ power. Thus, by Eq~\ref{eq:regularized_objective} we have $U_{PR}(X)=3-3\lambda$ and $U_{PR}(Y)=2$, so for $\lambda>\frac{1}{3}$ we prefer $Y$. See Table ~\ref{Tab:LargerAttackDefenseGame} for a larger variant of the game with a nontrivial Pareto frontier. Figure ~\ref{fig:attack_defense_action_values} shows the value of each action as a function of $\lambda$ assuming the coplayer is on-policy (i.e. doesn't play F).

It is important for the regularized objective to apply at every state, even those unreachable on policy.  A naive approach to penalizing power would be to only penalize the 1-step adversarial power over the agent in the initial state, that is, only aiming to maximize $U_i(\pi \vert s_0)$ where $s_0$ is the initial state. However, such a measure has a fundamental flaw, in that once an agent deviates from the usual strategy, there is no longer any incentive to regulate power and thus our agent would \emph{gain trust in potentially adversarial coplayers}.  

For instance, suppose the optimal power regularized policy were to work independently instead of forming an assembly line.  Once an agent deviates, the system could be in some state $s$ not reachable on-policy, only reachable via an adversarial deviation $a$.  The only way behavior in state $s$ influences the utility at the initial state $U_i(\pi \vert s_0)$ is through the adversarial action term of the power regularization $U_i^{\text{power}}(\pi \vert s_0)$. Since this term increases when task reward increases, after any deviation the policy will no longer regulate power.  Thus once one agent fails, all agents would revert to forming brittle and power-concentrating assembly lines.  This is the opposite of the desired behavior: we would take a failure of an agent as an indication that they should be entrusted with more power because our model does not allow them to deviate again.

Luckily, the state conditioned regularization we propose is a simple fix. Rather than regularizing for power just at the first state, we regularize power via optimizing Eq ~\ref{eq:regularized_objective} at all $s$. Thus even after an agent fails others will still continue to regularize for power.

\section{Existence of Equilibria}
In the standard formulation of Markov Games, the existence of an equilibrium solution is guaranteed by Nash's Theorem, which shows that every finite game has a mixed Nash equilibrium.  However, once we regularize for power, Nash's theorem no longer applies because the payoff becomes a function of the strategy.  

Given our power-regularized objective, we can define notions of best response and equilibrium similar to the standard formulations.
\begin{definition}
We say that a policy $\pi_i$ is a $\lambda$-power regularized best response to the policies $\pi_{-i}$, notated as $\pi_i \in$ $PRBR_{\lambda}(\pi_{-i})$, if it achieves the optimal trade off between task reward and power minimization in every state.  That is, for all $s$ we have:
\[
\pi_i \in \argmax_{\pi_i' \in \Pi_i}\{U_i(\pi_i'; \pi_{-i} \vert s)\} = PRBR_{\lambda}(\pi_{-i} \vert s).
\]
\end{definition}

Next, we define Power Regularizing Equilibrium (PRE) to be a fixed point of the power regularized best response function and then prove they are guaranteed to exist in any game.

\begin{definition}[$\lambda$-Power Regularizing Equilibrium]
A $\lambda$-Power Regularizing Equilibrium (PRE) is a policy tuple $\pi$ such that all policies are power regularized best responses to the others.  That is, in every state $s$, for all $i$, we have $\pi_i \in PRBR_{\lambda}(\pi_{-i} \vert s).$
\end{definition}

\begin{theorem} \label{theorem:pre_existence}
Let $G$ be a finite, discrete Markov Game, then a $\lambda$-power regularizing equilibrium exists for any $\lambda$.
\end{theorem}

Intuitively, we prove this by constructing another game, which we call the $p$-adversarial game of $G$, to which Nash's theorem can be applied, and show that Nash equilibria in this modified game correspond to power regularizing equilibria in the original game.  The basic idea of this game is to add adversarial players that perturb co-players' actions adversarially toward the ego agent with probability $p$.  We define the $p$-adversarial game formally below and depict the extensive form of a 1-timestep game in Figure ~\ref{fig:efg_for_existence_proof}. 

\begin{definition}[p-Adversarial Game of $G$ at $s$] \label{def:p_adversarial_game}
The p-adversarial game of $G$ at state $s$ adds adversarial agents $\pi_j^{A^*i}$ for each player $i$ and co-player $j$. These agents are randomly given control of $i$'s co-players to minimize $i$'s return. The game starts at $s$. Nature randomly decides with probability $p$ to let the adversary will take control in this episode.  Nature uniformly randomly selects an ego-agent $i$ which will be the only agent to be rewarded in the game, and uniformly randomly selects a time step on which the adversary will take control, if the adversary gets control this episode. At that time step Nature will, choose a co-player $j$ to be replaced by $\pi_j^{A^*i}$ for one step. Rewards for $i$ are calculated as normal.
\end{definition}

The following theorem establishes a correspondence between the best response in the p-adversarial game of $G$ and the power regularized best response in the original game.  We will use this correspondence to prove Theorem ~\ref{theorem:pre_existence}.

\begin{theorem} \label{theorem:adv_game_equivelence}
Consider an agent $i$ in a Markov game $G$ with time horizon $T$, and an arbitrary state s. The utility of policies in the p-adversarial game at state s, for $p= \lambda$ is equivalent to the corresponding policies in the original game.  That is, for all $\pi$ in the original Markov game, we have:
$U_i^{\text{p-Adv}}(\pi_i; \pi_{-i} \vert s) = U_i(\pi_i; \pi_{-i} \vert s).$
\end{theorem}

The proof idea is to, for a p-adversarial game on a fixed state s, inductively argue for the equivalence starting from the last step of the game.  At each step, the set of $\pi_j^{A^*i}$ for all co-players $j$ can be seen as a way to compute $i$'s power regularization term.  Throughout the proof we use state-based rewards to simplify the notation, which otherwise does not effect the main structure of the proof.

\begin{proof}


\textbf{Base Case.} Without loss of generality, assume that all trajectories end in a single state where agents' decisions affect nothing.  Such a state can be added without changing the power or utility of any trajectory. At this state all policies are equally valued, so the base case holds trivially. 


\textbf{Inductive Step.} Assume that, at any state $s'$ reachable at time $t-1$ from the end, $U_i^{\text{p-Adv}}(\pi_i; \pi_{-i} \vert s') = U_i(\pi_i; \pi_{-i} \vert s')$. Our goal is to show this equivalence also holds for states reachable at time $t$. Expanding out the definition of the p-Adversarial game, we have:
\begin{alignat*}{3}
    & U_i^{\text{p-Adv}}(\pi_i'&&; \pi_{-i} \vert s)
    \\=& R_i^{\text{task}}(s,\pi_{-i}) &&+ (1-  \frac{\lambda}{T})\E
    [U_i^{\text{p-Adv}}(\pi_i'; \pi_{-i} \vert s')] \\& &&+   \frac{\lambda}{T}\E
    [U_i^{\text{task}}(\pi_i'; \pi_{-i} \vert s_{Adv}')]
    \\=& R_i^{\text{task}}(s) &&+ (1-  \frac{\lambda}{T})\E\limits
    [U_i(\pi_i'; \pi_{-i} \vert s')] \\& &&+   \frac{\lambda}{T}\E
    [U_i^{\text{task}}(\pi_i'; \pi_{-i} \vert s_{Adv}')] \\
\end{alignat*}
where $s' \sim T(\pi_i';\pi_{-i})$, $s_{Adv}' \sim T(\pi_i';\pi_j^{A^*i};\pi_{-\{i,j\}})$, and $U_i^{\text{p-Adv}}(\pi \vert s)$ is the utility of agent $i$ given policies $\pi$ starting at state $s$ of the $p$ Adversarial game before the adversary has taken control.  The first line above follows from the definition of the $p$ adversarial game and the second line follows from the inductive hypothesis. We can continue by expanding out the definitions and rearranging terms:
\begin{alignat*}{3}
& R_i^{\text{task}}(s) &&+ (1-  \frac{\lambda}{T})\E
    [U_i^{task}(\pi_i'; \pi_{-i} \vert s')] \\
    &+ \lambda (1-  \frac{\lambda}{T})\E
    [U_i^{\text{power}}(\pi_i'; \pi_{-i} \vert s')] &&+   \frac{\lambda}{T}\E
    [U_i^{\text{task}}(\pi_i'; \pi_{-i} \vert s_{Adv}')]
    \\=& R_i^{\text{task}}(s) &&+ \E
    [U_i^{\text{task}}(\pi_i'; \pi_{-i} \vert s')] \\
    & &&+ \lambda (1-  \frac{\lambda}{T})\E
    [U_i^{\text{power}}(\pi_i'; \pi_{-i} \vert s')] \\
    & -   \frac{\lambda}{T}\Bigl(\E
    [U_i^{\text{task}}(\pi_i'; \pi_{-i} \vert s')] 
    &&-\E
    [U_i(^{\text{task}}\pi_i'; \pi_{-i} \vert s_{Adv}')]\Bigr) 
    \\=& U_i^{\text{task}}(\pi \vert s) &&+ \lambda U_i^{\text{power}}(\pi \vert s)\\
\end{alignat*}
where the final line follows from the definition of task and power utility. Thus, the value of policies in the p-Adversarial game at state $s$ time step $t$ is equivalent to the power-regularized value..

By induction, the equivalence holds for states reachable at any time step. Thus, we have the desired equivalence $
U_i^{\text{p-Adv}}(\pi_i; \pi_{-i} \vert s) = U_i(\pi_i; \pi_{-i} \vert s).$
\end{proof}

Given the equivalence between the standard best response in the p-adversarial game and the power-regularized best response in the original game, we can return to our task of proving Theorem \ref{theorem:pre_existence}, to show that power regularizing equilibria of $G$ always exist.

\begin{proof}
Note that we can construct a tuple of policies $\pi$ which are a local Nash equilibrium at $s$ in the $p$-adversarial game at $s$ by standard backwards induction argument -- noting first that this can be done in the terminal states, and then noting that it can then be done at each of the prior states backwards by induction. 

Consider the policy $\pi_i$ of an arbitrary player $i$, by Theorem \ref{theorem:adv_game_equivelence}, $\pi_i$ must be a power regularized best response in $G$ at state $s$.  Since we assumed $i$ and $s$ to be arbitrary, this applies to all $i$ and $s$.  Thus the tuple $\pi_i$ represents a power regularizing equilibrium of $G$ at all states $s$ by definition. 
\end{proof}

We have shown that the power regularizing equilibria we seek do, in fact, exist, but moreover, Theorem \ref{theorem:adv_game_equivelence} gives some idea about a method to actually obtain them. We can find policies in power-regularizing equilibrium by finding policies in Nash equilibrium in the p-adversarial game of $G$.  This intuition is the motivation behind one of our methods, Sample Based Power Regularization, which we introduce in Section \ref{section:sbpr} and empirically evaluate in Section \ref{sec:experiments}.

\section{Methods}
We introduce two methods for power regularization, Sample Based Power Regularization (SBPR) and Power Regularization via Intrinsic Motivation (PRIM). SBPR is inspired by the p-Adversarial Game formulation introduced in Definition ~\ref{def:p_adversarial_game}; it injects adversarial data during training by perturbing actions at each step with probability $p=\lambda$. PRIM trains agents directly on the power regularized objective, interpreting the power penalty as intrinsic motivation.

Agents do not share weights, but we train them together offline. Our theory and methods are amenable to be combined with approaches from ad-hoc team play~\cite{AAAI10-adhoc,barrett2011empirical} and zero-shot coordination~\cite{hu2020other,treutlein2021new}, though these domains bring with them their own challenges, so we leave it to future work to generalize this approach. 

We train each agent using Proximal Policy Optimization (PPO). The neural networks parameterizing the policy consist of several convolutional layers, fully connected layers, a categorical output head for the actor, and a linear layer value function output head.

\subsection{Sample-Based Power Regularization (SBPR)} \label{section:sbpr}

SBPR directly injects adversarial rollouts into the training data, playing the p-Adversarial Game introduced in Definition ~\ref{def:p_adversarial_game}. At the beginning of every rollout, we pick an agent $i$ to be the ego agent and another agent $j$ to be the adversary. At every timestep, with probability $p$ (independent of previous timesteps) we perturb agent $j$'s action adversarially to $i$. Only agent $i$ receives the rollout to train on. Intuitively, this can be seen as training on the $p$-Adversarial Game at random states $s$.



SBPR has the advantage of simplicity but may not regularize for power successfully if we want $~\lambda$ to be very small. This is because we won't see enough deviation examples per batch, so the gradient signal is very high variance, which is reflected in our experiments.

\begin{algorithm}
\caption{Sample Based Power Regularization (SBPR)}\label{sbpr}
\begin{algorithmic}[1]
\Procedure{trainAgent}{$\pi_i$, $\pi_j$, $\hat{\pi}_j$, $V_i$, $p$}
\State Collect trajectories for agent $i$: at each timestep, always use $\pi_i$, and with probability $p$ use $\hat{\pi}_j$, otherwise use $\pi_j$.
\State Use PPO or other RL algorithm to update $\pi_i$ and $V_i$.
\EndProcedure
\Procedure{trainAdversarialAgent}{$\pi_i$, $\hat{\pi}_j$, $V_i$}
\State Collect trajectories for adversarial agent $j$: $\hat{\pi}_j$ has one step to act, and $r_i=V_i(s)-r(s,\pi_i(s),\hat{\pi}_j(s))-\gamma V_i(s')$.
\State Use PPO or other RL algorithm to update $\hat{\pi}_j$.
\EndProcedure
\Procedure{SBPR}{$p$}
\State Initialize $\pi_i$, $\pi_j$, $\hat{\pi}_i$, $\hat{\pi}_j$, $V_i$, $V_j$ arbitrarily.
\Loop
\State $\Call{trainAgent}{\pi_i,\pi_j,\hat{\pi}_j, p}$
\State $\Call{trainAgent}{\pi_j,\pi_i,\hat{\pi}_i, p}$
\State $\Call{trainAdversarialAgent}{\pi_j,\hat{\pi}_i, V_j}$
\State $\Call{trainAdversarialAgent}{\pi_i,\hat{\pi}_j, V_i}$
\EndLoop
\EndProcedure
\end{algorithmic}
\end{algorithm}

\subsection{Power Regularization via Intrinsic Motivation (PRIM)}

PRIM adds a per-step intrinsic motivation reward term that penalizes power on the agent. Each agent's reward function is
\[R_i^{PRIM}(s,a,\pi)=R_i^{task}(s,a)+\lambda R_i^{power}(s,\pi)\]

which is precisely the power regularization objective of Eq~\ref{eq:regularized_objective}. Rather than probabilistically considering the effect of an adversary like SBPR, PRIM considers it at every step but downweights the effect according to $\lambda$, which reduces variance.

Crucially, finding $j$'s adversarial action for $i$ to compute adversarial power requires counterfactuals from a resettable simulator. In the future one could try to learn the simulator instead.


\begin{algorithm}
\caption{Power Regularization via Intrinsic Motivation (PRIM)}\label{prim}
\begin{algorithmic}[1]
\Procedure{ComputePower}{$\pi_i$, $\pi_j$, $\hat{\pi}_j$, $V_i$}
\State $s' \sim T(\cdot | s, \pi_i(s), \pi_j(s))$
\State $r \gets r(s, \pi_i(s), \pi_j(s)) + \gamma V_i(s')$
\State $s_{adv}' \sim T(\cdot | s, \pi_i(s), \hat{\pi}_j(s))$
\State $r_{adv} \gets r(s, \pi_i(s), \hat{\pi}_j(s)) + \gamma V_i(s_{adv}')$
\State $power \gets r - r_{adv}$
\State \Return $power$
\EndProcedure
\Procedure{trainAgent}{$\pi_i$, $\pi_j$, $\hat{\pi}_j$, $V_i$, $\lambda$}
\State Collect trajectories for agent $i$ with reward $r_{task}+\lambda r_{power}$.
\State Use PPO or other RL algorithm to update $\pi_i$ and $V_i$.
\EndProcedure
\Procedure{trainAdversarialAgent}{$\pi_i$, $\hat{\pi}_j$, $V_i$}
\State Collect trajectories for adversarial agent $j$: $\hat{\pi}_j$ has one step to act, and $r_i=V_i(s)-r(s,\pi_i(s),\hat{\pi}_j(s))-\gamma V_i(s')$.
\State Use PPO or other RL algorithm to update $\hat{\pi}_j$.
\EndProcedure
\Procedure{PRIM}{$\lambda$}
\State Initialize $\pi_i$, $\pi_j$, $\hat{\pi}_i$, $\hat{\pi}_j$, $V_i$, $V_j$ arbitrarily.
\Loop
\State $\pi_i, V_i \gets \Call{trainAgent}{\pi_i,\pi_j,\hat{\pi}_j, V_i, \lambda}$
\State $\pi_j, V_j \gets \Call{trainAgent}{\pi_j,\pi_i,\hat{\pi}_i, V_j, \lambda}$
\State $\hat{\pi}_i \gets \Call{trainAdversarialAgent}{\pi_j,\hat{\pi}_i, V_j}$
\State $\hat{\pi}_j \gets \Call{trainAdversarialAgent}{\pi_i,\hat{\pi}_j, V_i}$
\EndLoop
\EndProcedure
\end{algorithmic}
\end{algorithm}

\begin{figure*}[h]
\begin{subfigure}{.3\textwidth}
\centering
  \includegraphics[width=\linewidth]{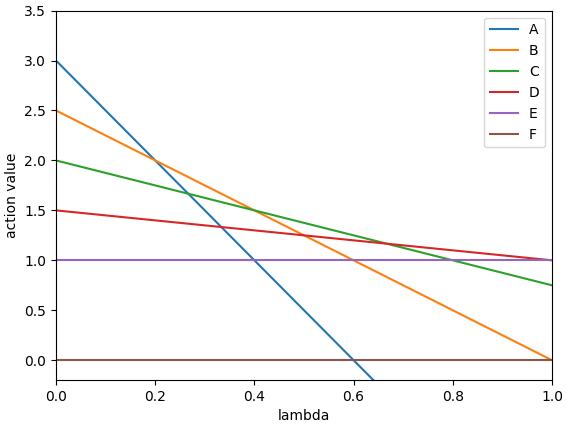}
  \caption{Attack-Defense Game Action Payoffs}
  \label{fig:attack_defense_action_values}
  \Description{Attack-Defense Matrix Game Action Payoffs}
\end{subfigure}
\begin{subfigure}{.3\textwidth}
  \centering
  \includegraphics[width=\linewidth]{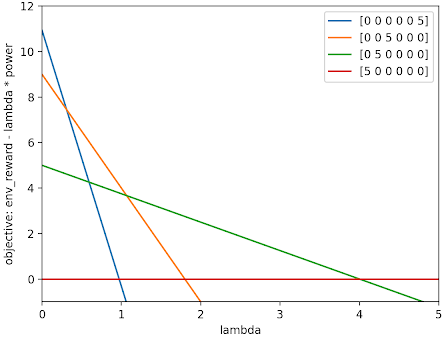}
  \caption{Coin Division Game Action Payoffs}
  \label{fig:coins_mean_action_values}
\end{subfigure}
\begin{subfigure}{.3\textwidth}
  \centering
  \includegraphics[width=\linewidth]{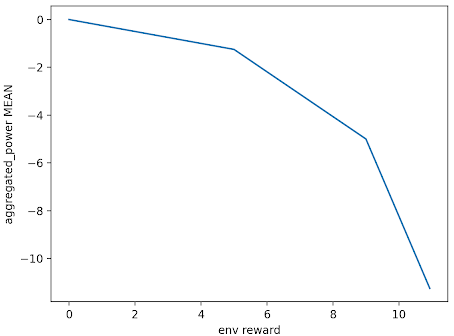}
  \caption{Coin Division Game Pareto Frontier}
  \label{fig:coins_mean_pareto}
\end{subfigure}
\Description{Action Payoffs in the Attack Defense and Coins Division Games}
\caption{Power-regularized objective values achieved by different actions in small environments.}
\end{figure*}

\subsection{Optimizations}
On their own, both PRIM and SBPR can be slow or unstable, so we introduce a number of optimizations. 

\textbf{Monte Carlo Estimates of Adversarial Power.} 
When computing adversarial power for PRIM, we use the instantiated actions in each rollout rather than the full policies $\pi$, effectively a Monte Carlo estimate. This is because PPO looks at the advantages of actions across the batch, so we need to be able to tell which actions incur more or less power. Policies don't give us this information because they are constant throughout a batch; we get no differential information. As a bonus, using actions gives a speedup by a factor of the joint co-player action space.

\textbf{Learning the Adversary.} 
Both PRIM and SBPR require finding the co-player $j$'s action that minimizes the ego agent $i$'s reward. For environments with large or continuous action spaces, conducting an exhaustive search may be intractable, so we learn the reward-minimizing action: the adversarial co-player $j$ for ego agent $i$ is trained to minimize $i$'s return:$U_{j,adv}^i(s,a_j)=-R_i(s,a)-\E[U_i(T(s,a))]$
where $a=\{a_j,\pi_{-j}(s)\}$. Each agent must maintain an adversarial model of each co-player.

\textbf{Using the Value Function to Approximate Return from Rollouts.} \label{section:vf_approx}
Both PRIM and SBPR require computing the value of states after an adversarial co-player has acted. The benefit of this trick is two-fold: one, it reduces variance because rollouts can be extremely noisy, especially in the beginning of training, and two, it speeds up runtime significantly.

\textbf{Domain Randomization (DR).} 
DR is useful for speeding up and stabilizing convergence in both methods. Overcooked is a highly sequential environment, requiring a long string of actions to receive a reward, so it is helpful to train starting from random states and learn the optimal policy backwards. Furthermore, crucially for PRIM, DR enables accurate value estimates of states that are off-policy and thus normally not visited. This allows agents to learn how to recover from adversarial deviations and update their value estimates of such states accordingly.

\textbf{Normalization for the Adversary.}
The adversary's objective is highly dependent on the starting state because it only gets to act for one timestep, thus the value is high variance which is only worsened by DR. We reduce variance by normalizing the adversary's reward by subtracting the value estimate of the starting state:
$$U_{j,adv}^i(s,a_j)=U_i(s)-R_i(s,a)-\E[U_i(T(s,a))]$$ where $a=\{a_j,\pi_{-j}(s)\}$.

\section{Experiments} \label{sec:experiments}

\begin{figure}
\begin{subfigure}{.49\columnwidth}
  \centering
  \includegraphics[width=0.299\linewidth]{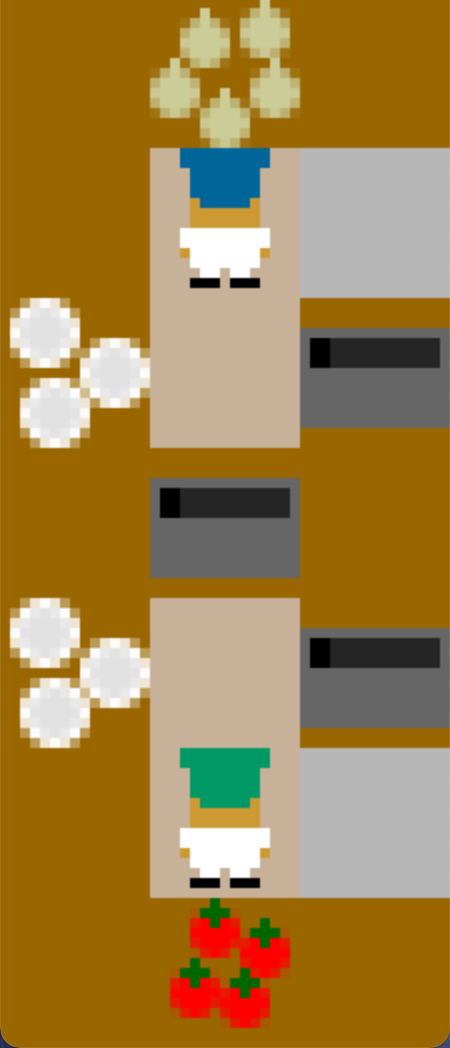}
  \caption{Starting State}
  \label{fig:overcooked_closepotfarpot_layout}
\end{subfigure}
\begin{subfigure}{.49\columnwidth}
  \centering
  \includegraphics[width=0.3\linewidth]{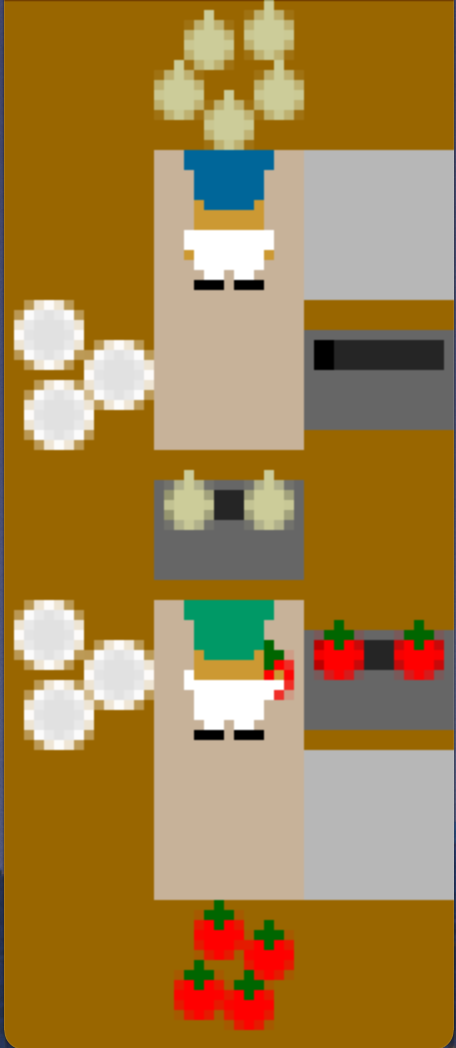}
  \caption{High Power Timestep}
  \label{fig:overcooked_high_power_timestep}
\end{subfigure}
\caption{Overcooked Close-Pot-Far-Pot. Agents can use the shared middle pot or their private pots. Using the middle pot is faster but incurs high power (see (b)) where one agent can mess up the other's work by putting in a wrong ingredient.}
\end{figure}

\begin{figure*}
\begin{subfigure}{.33\textwidth}
  \centering
  \includegraphics[width=\linewidth]{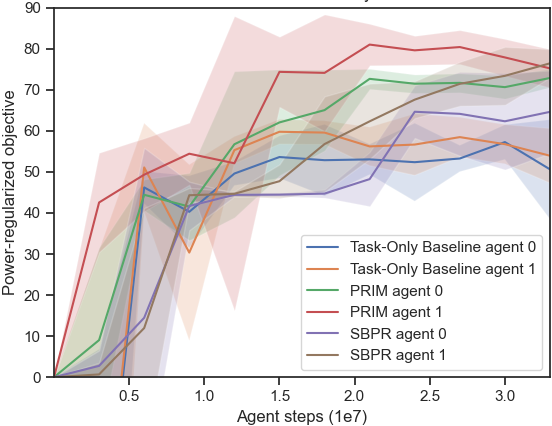}
  \caption{PRIM vs SBPR vs Task-Only Baseline}
  \label{fig:prim_vs_sbpr_vs_baseline}
\end{subfigure}
\begin{subfigure}{.33\textwidth}
  \centering
  \includegraphics[width=\linewidth]{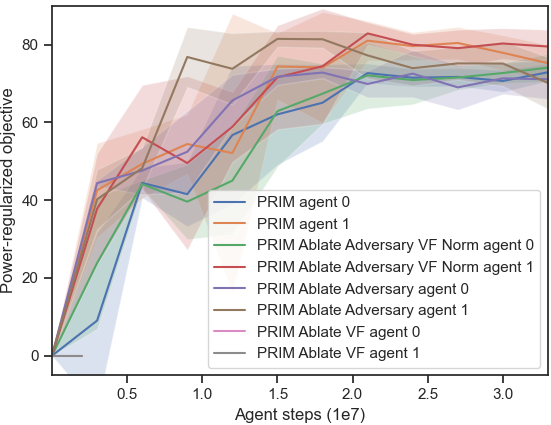}
  \caption{PRIM Ablations}
  \label{fig:prim_ablations}
\end{subfigure}
\begin{subfigure}{.33\textwidth}
  \centering
  \includegraphics[width=\linewidth]{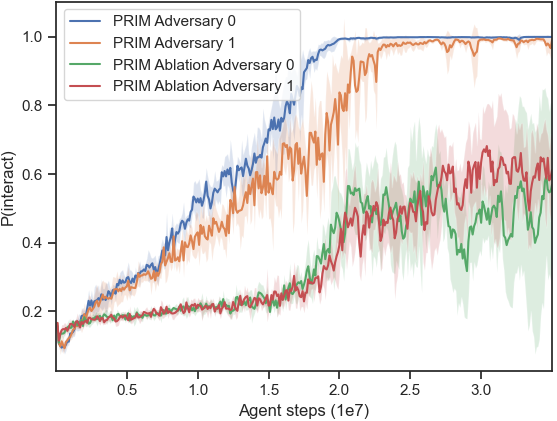}
  \caption{Adversary Policy Convergence}
  \label{fig:prim_ablate_adv_norm_interact_prob}
\end{subfigure}
\Description{Experiments in Overcooked Close-Pot-Far-Pot}
\caption{Experimental Results in Overcooked Close-Pot-Far-Pot. Error bars are standard deviations over 5 trials.}
\end{figure*}

\begin{table*}
  \caption{End-of-training metrics in Overcooked Close-Pot-Far-Pot. Error values are standard deviations over 5 trials.}
  \label{tab:results}
  \begin{tabular}{llllll}\toprule
    \textit{Name} & \textit{Task reward} & \textit{Power on Agent 0} & \textit{Power on Agent 1} & \textit{PR Objective Agent 0} & \textit{PR Objective Agent 1} \\ \midrule
    Task-only baseline & 104.9 $\pm$ 0.2 & 217.1 $\pm$ 47.9 & 203.8 $\pm$ 26.1 & 50.6 $\pm$ 12.1 & 53.9 $\pm$ 6.6 \\
    SBPR & 94.2 $\pm$ 0.0 & 118.4 $\pm$ 40.3 & 71.1 $\pm$ 13.8 & 64.6 $\pm$ 10.1 & 76.4 $\pm$ 3.4 \\
    PRIM & 94.4 $\pm$ 0.1 & 86.0 $\pm$ 9.2 & 76.6 $\pm$ 18.6 & 72.9 $\pm$ 2.2 & 75.2 $\pm$ 4.7 \\
    PRIM ablate adversary norm & 94.3 $\pm$ 0.1 & 80.8 $\pm$ 14.5 & 59.1 $\pm$ 17.2 & 74.0 $\pm$ 3.5 & 79.5 $\pm$ 4.3 \\
    PRIM ablate adversary & 94.5 $\pm$ 0.2 & 93.6 $\pm$ 20.3 & 97.4 $\pm$ 27.1 & 71.1 $\pm$ 5.1 & 70.1 $\pm$ 6.6 \\
    PRIM ablate VF & 0.0 $\pm$ 0.0 & 0.0 $\pm$ 0.0 & 0.0 $\pm$ 0.0 & 0.0 $\pm$ 0.0 & 0.0 $\pm$ 0.0 \\ \bottomrule
  \end{tabular}
\end{table*}

We first validate our methods in small environments where we can compute the optimal actions and then move to larger environments. 

\subsection{Small Environments}

We evaluate in the larger version of the Attack Defense Game (payoff matrix given in Table ~\ref{Tab:LargerAttackDefenseGame} and power-regularized objective values per action in Figure ~\ref{fig:attack_defense_action_values}) and another environment called the Coin Division game. There are four agents, one "divider" agent (P0), and three "accepter" agents (P1, P2, and P3). There are six bins with the following assignment of agents to bins: ([], [P0], [P0,P1], [P0,P2], [P1,P2], [P1,P2,P3]). The divider agent must allocate five coins amongst the bins. For each bin, the agents assigned to that bin have the option of accepting or rejecting. If everyone accepts, everyone takes home the number of coins assigned to that bin times the number of agents assigned to that bin. If one or more agents reject, the coins assigned to that bin are destroyed. We consider the divider agent's optimal policy and assume that all accepter agents always accept $90\%$ of the time (per bin).

See Figure ~\ref{fig:coins_mean_action_values} for the power-regularized objective values of each action (omitting actions which are strictly dominated) and Figure ~\ref{fig:coins_mean_pareto} for the corresponding Pareo frontier. Both PRIM and SBPR achieve the optimal actions for values of $\lambda$ sampled in the range 0 to 1.

\begin{figure*}
\begin{subfigure}{.33\textwidth}
  \centering
  \includegraphics[width=\linewidth]{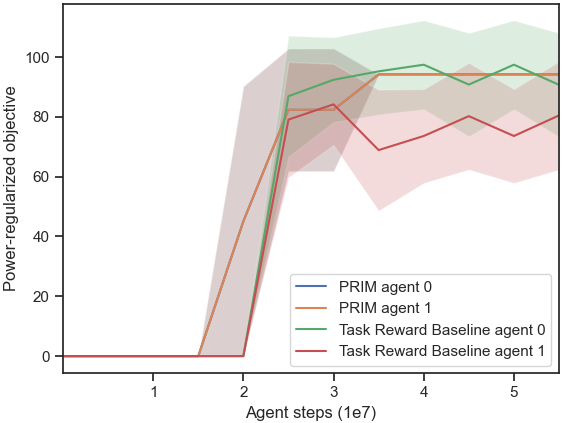}
  \caption{PRIM vs Baseline (Explosion)}
  \label{fig:power_reg_not_free_explosion}
\end{subfigure}
\begin{subfigure}{.33\textwidth}
  \centering
  \includegraphics[width=\linewidth]{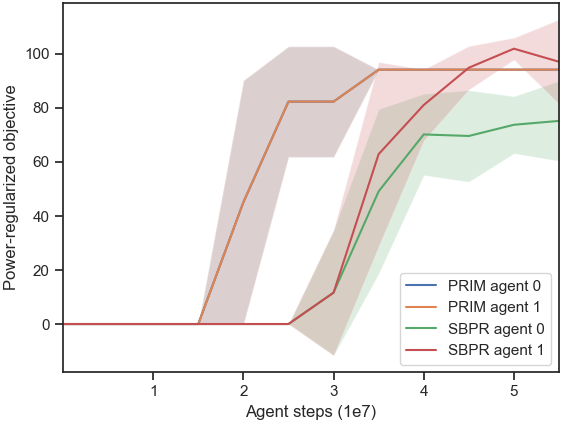}
  \caption{PRIM vs SBPR (Explosion)}
  \label{fig:prim_vs_sbpr_explosion}
\end{subfigure}
\begin{subfigure}{.33\textwidth}
  \centering
  \includegraphics[width=\linewidth]{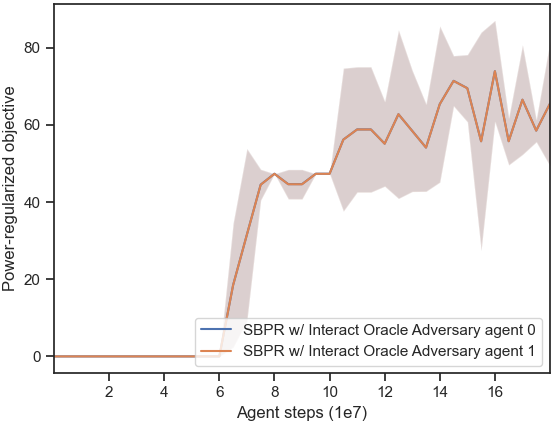}
  \caption{PRIM vs SBPR (Explosion)}
  \label{fig:sbpr_interact_oracle_explosion}
\end{subfigure}
\Description{Explosion results}
\caption{Comparison of PRIM, SBPR, and Task-Only Baseline in Overcooked Explosion with $\lambda=0.0001$. In some runs only one agent is visible because the plots coincide completely; the powers incurred are too small to be distinguishable after multiplying by $\lambda$. Error bars are standard deviations over 5 trials except for SBPR interact oracle which only has 3 trials.}
\end{figure*}

\begin{table*}
  \caption{Experimental Results in Explosion Environment. Error values are standard deviations over 5 trials except for SBPR interact oracle which only has 3 trials.}
  \label{tab:results_explosion}
  \begin{tabular}{llllll}\toprule
    \textit{Name} & \textit{Task reward} & \textit{Power on Agent 0} & \textit{Power on Agent 1} & \textit{PR Objective Agent 0} & \textit{PR Objective Agent 1} \\ \midrule
    Task-only baseline & 104.7 $\pm$ 0.4 & 139949.5 $\pm$ 171647.1 & 242363.6 $\pm$ 182484.0 & 90.7 $\pm$ 17.3 & 80.5 $\pm$ 18.1 \\
    SBPR & 105.0 $\pm$ 0.0 & 297351.0 $\pm$ 148585.0 & 78369.8 $\pm$ 156414.3 & 75.2 $\pm$ 14.8 & 97.1 $\pm$ 15.7 \\
    SBPR INTERACT adversary & 65.5 $\pm$ 16.2 & 174.0 $\pm$ 176.0 & 217.8 $\pm$ 212.6 & 65.5 $\pm$ 16.2 & 65.5 $\pm$ 16.2 \\ 
    PRIM & 94.2 $\pm$ 0.0 & 138.4 $\pm$ 2.7 & 132.3 $\pm$ 40.5 & 94.2 $\pm$ 2.2 & 94.2 $\pm$ 0.0 \\ \bottomrule
  \end{tabular}
\end{table*}

\subsection{Overcooked: Close-Pot-Far-Pot}

We evaluate both SBPR and PRIM in Overcooked, a $2$ player gridworld game where the objective is to prepare and deliver soups according to given recipes. Recipes may call for two types of ingredients, tomatoes and onions. Agents must collect and place all ingredients in a pot one at a time, cook the soup, grab a plate, place the finished soup onto the plate, and finally deliver the soup. 

The action space is $\{\texttt{N}$, $\texttt{E}$, $\texttt{S}$, $\texttt{W}$, $\texttt{STAY}$, $\texttt{INTERACT}\}$. Depending on where the agent is facing, $\texttt{INTERACT}$ can mean pick up an ingredient, place an ingredient into a pot, start cooking, pick up a dish, place soup onto the dish, or deliver a soup. It's impossible to remove ingredients from a pot once they are placed.

We design a layout "Close-Pot-Far-Pot" with two recipes, 3 tomatoes or 3 onions, each giving $R$ reward. The top agent can only access onions and the bottom agent can only access tomatoes. Each agent can access two pots, one shared in the center and the other is private, inaccessible to the other agent, but further. The agents share a reward function and a trajectory is $T$ steps.

In our experiments we set $T=105$ and $R=20$. An assembly line (strategy 1) using the middle pot can produce $7$ soups, one agent independently using the middle pot and the other using their private pot can produce $9$ soups (strategy 2), and both agents independently using their private pots can produce $8$ soups (strategy 3).

Strategy 2 maximizes task reward but incurs high adversarial power: as shown in Figure~\ref{fig:overcooked_high_power_timestep}, the tomato agent can mess up the onion agent's soup by putting in a wrong ingredient, leading to the onion agent making fewer soups. The state depicted in Figure~\ref{fig:overcooked_high_power_timestep} is on-policy since the tomato agent must move up before turning right to face its private pot to place its tomato there.

We compare the performance of our methods to the task-only baseline. We compute ground truth power through an exhaustive search for the return-minimizing action and conduct full rollouts to evaluate resulting states. This is extremely slow so we only calculate it once every several hundred training iterations. In general rollouts are high variance so multiple trials should be performed, but since our agents converge towards deterministic policies in our environments, we simply determinize the policies when rolling out.

For $\lambda=0.25$, Figure ~\ref{fig:prim_vs_sbpr_vs_baseline} shows that PRIM outperforms the baseline of optimizing for just the task reward. PRIM also performs better than SBPR for one agent due to its inherently lower variance training data which makes the learning problem easier. 

Next we ran a series of ablation experiments to better understand PRIM, shown in Figure ~\ref{fig:prim_ablations}. Ablating the learned adversary and instead conducting an exhaustive search over the action space did not make much difference on the objective value achieved. This is expected; the goal of learning the adversary is simply to speed up the power computation: rather than iterating over the action space, we pay a "fixed cost" to train and query the adversary. This is is necessary in environments with large action spaces.

Ablating normalization for the adversary's objective did not significantly change the objective value achieved, but it did hurt the adversary's convergence. Figure ~\ref{fig:prim_ablate_adv_norm_interact_prob} depicts the poor convergence for the state in Figure ~\ref{fig:overcooked_high_power_timestep} where the optimal action is $\texttt{INTERACT}$.

Finally we ablated the use of value function to approximate the return from a rollout. We ran this experiment for the same amount of time as the other experiments, but it was so slow it was only able to finish 2e6 agent steps and achieved 0 on the task reward.  We summarize the results of our experiments in Table ~\ref{tab:results}.

\subsection{Overcooked: Explosion}

In the Close-Pot-Far-Pot layout, an adversarial deviation does not have large consequences, but power regularization may be more useful in high stakes events. We create a variant of Close-Pot-Far-Pot called Explosion where we interpret the ingredients as chemicals, the pots as test tubes, and the recipes as chemical formulas. If unlike chemicals are mixed together, a dangerous chemical reaction causes an explosion which incurs an immediate penalty of $P=-100,000$. 

Figure ~\ref{fig:power_reg_not_free_explosion} compares the task reward only baseline to PRIM with $\lambda=0.0001$. Note that the blue line is hidden beneath the orange line. PRIM converges to very low variance while the baseline has high variance.  This is due in large part to the fact that agents 0 and 1 may switch roles in who uses the shared pot so either agent may incur the large power penalty.

Now we examine SBPR's performance (see Figure ~\ref{fig:prim_vs_sbpr_explosion}). We expected SBPR to fail since the probability of a deviation $p=0.0001$ is so low yet the explosion penalty $P=-100,000$ is so high, but the observed performance was better than expected. However, a significant amount of hyperparameter tuning was necessary: we adjusted the PPO clip param and maximum grad norm down to $0.1$ and lengthened the entropy schedule. Depending on the particular hyperparameter values, the agents would either fail to optimize for power at all or would converge on an assembly line that avoids the explosion risk (but is suboptimal to PRIM's solution).

As shown in Figure ~\ref{fig:sbpr_interact_oracle_explosion}, SBPR relies on the adversary not yet converging at the beginning because this allows the agents to solve enough of the exploration problem before consistently incurring the penalty. Replacing the adversary with an agent that always plays INTERACT (an \emph{interact oracle}) causes SBPR to fail.

We summarize the Explosion results in Table ~\ref{tab:results_explosion}. PRIM is the only method that avoids incurring catastrophically high power at the cost of a bit of task reward.

\section{Conclusion}
We defined a notion of power amenable to optimization and showed that equilibria always exist when agents regularize for power. Next, we presented two algorithms, Sample Based Power Regularization (SBPR) and Power Regularization via Intrinsic Motivation (PRIM). We validate our methods in a series of small environments and in two variants of Overcooked, showing that both methods guide agents toward lower power behavior. SBPR is simpler but PRIM is better able to handle very low values of $\lambda$.

There are many avenues for future work, including exploring different definitions of power (empirically and philosophically) and modeling multiple timestep deviations. Our theoretical results hold for general-sum games but we have not explored general-sum games empirically.

\begin{acks}
We are grateful for insightful conversations from the members of the Center for Human-Compatible AI, including Micah Carroll, Niklas Lauffer, Adam Gleave, Daniel Filan, Lawrence Chan, and Sam Toyer, as well as Derek Yen from MIT. We are also grateful for funding of this work as a gift from the Berkeley Existential Risk Initiative.
\end{acks}



\bibliographystyle{ACM-Reference-Format} 
\balance
\bibliography{sample}



    


\end{document}